\documentclass[11pt]{article}
\usepackage{acl2014}
\usepackage{times,url,color,float}
\usepackage{latexsym}
\usepackage{amsmath,amsfonts,amssymb}
\usepackage{algorithm,algorithmic}
\usepackage{graphicx}
\usepackage{booktabs}
\usepackage{multirow}
\usepackage{tikz,tikz-qtree}
\usepackage{paralist}
\usepackage{subcaption}
\usepackage{mystuff-nlp}
\usepackage{gb4e}
\usepackage{enumitem}

\newcommand{\ontodata}{PDTB$\cap$Onto}
\newcommand{\modelname}{\textsc{disco2}}
\newcommand{\updownname}{Up-Down Composition}

\newcommand{\jacob}[1]{[\textcolor{purple}{#1 - JE}]}

\newcommand{\true}[1]{{#1}^{\ast}}
\newcommand{\rela}[1]{{\sc #1}}
\newcommand{\entity}[1]{\underline{#1}}
\newcommand{\uvec}[1]{\vec{u}_{#1}} 
\newcommand{\dvec}[1]{\vec{d}_{#1}} 
\newcommand{\fvec}[1]{\vec{\phi}_{#1}}
\newcommand{\alignset}[2]{\set{A}(#1,#2)} 
\newcommand{\uclass}[1]{\mat{A}_{#1}} 
\newcommand{\dclass}[1]{\mat{B}_{#1}} 
\newcommand{\fclass}[1]{\vec{\beta}_{#1}} 
\newcommand{\ucomp}{\mat{U}} 
\newcommand{\dcomp}{\mat{V}} 
\newcommand{\lchild}[1]{\ell(#1)} 
\newcommand{\rchild}[1]{r(#1)} 
\newcommand{\mytanh}[1]{\text{tanh}\left(#1\right)}
\newcommand{\example}[1]{\textit{#1}}
\renewcommand{\trans}[1]{\ensuremath{#1^{\top}}}

\usetikzlibrary{trees,positioning,arrows}

\newcommand{\vu}{\vec{u}}
\newcommand{\vul}{\vec{u}^{(\ell)}}
\newcommand{\vur}{\vec{u}^{(r)}}
\newcommand{\vum}{\vec{u}^{(m)}}
\newcommand{\vun}{\vec{u}^{(n)}}
\newcommand{\vd}{\vec{d}}
\newcommand{\vdl}{\vec{d}^{(\ell)}}
\newcommand{\vdr}{\vec{d}^{(r)}}
\newcommand{\vdm}{\vec{d}^{(m)}}
\newcommand{\vdn}{\vec{d}^{(n)}}
\tikzstyle{every node}=[font=\small]
\tikzstyle{up}=[edge from parent/.style={draw,latex-}]
\tikzstyle{down}=[edge from parent/.style={draw,-latex}]
\tikzstyle{nodraw}=[edge from parent/.style={}]

\title{One Vector is Not Enough:\\Entity-Augmented Distributional Semantics for Discourse Relations}

\author{Yangfeng Ji \and Jacob Eisenstein \\
  School of Interactive Computing \\
  Georgia Institute of Technology \\
  {\tt \{jiyfeng, jacobe\}@gatech.edu}}
\date{}

\begin{document}
\maketitle

\begin{abstract}
Discourse relations bind smaller linguistic units into coherent texts. However, automatically identifying discourse relations is difficult, because it requires understanding the semantics of the linked arguments. A more subtle challenge is that it is not enough to represent the meaning of each argument of a discourse relation, because the relation may depend on links between lower-level components, such as entity mentions. Our solution computes distributional meaning representations by composition up the syntactic parse tree. A key difference from previous work on compositional distributional semantics is that we also compute representations for entity mentions, using a novel downward compositional pass. Discourse relations are predicted from the distributional representations of the arguments, and also of their coreferent entity mentions. The resulting system obtains substantial improvements over the previous state-of-the-art in predicting implicit discourse relations in the Penn Discourse Treebank.
\end{abstract}



\section{Introduction}

The high-level organization of text can be characterized in terms of \textbf{discourse relations} between adjacent spans of text~\cite{knott1996data,mann1984discourse,webber1999discourse}. Identifying these relations has been shown to be relevant to tasks such as summarization~\cite{louis2010discourse,yoshida2014dependency}, sentiment analysis~\cite{somasundaran2009supervised}, and coherence evaluation~\cite{lin2011automatically}. While the Penn Discourse Treebank (PDTB) now provides a large dataset annotated for discourse  relations~\cite{prasad2008penn}, the automatic identification of implicit relations is a difficult task, with state-of-the-art performance at roughly 40\%~\cite{lin2009recognizing}.


\begin{figure}
\vspace{-5pt}
\begin{subfigure}{0.48\textwidth}
\centering
\begin{tikzpicture}[level distance=22pt,sibling distance=24pt]
  \node (arg0) [rectangle,draw=gray] {$\vul_0$}
  child[up] {node {Bob}}
  child[up] {node {$\vul_1$}
    child[up] {node {gave}}
    child[up] {node {$\vul_2$}
      child[up] {node {Tina}}
      child[up] {node {$\vul_3$}
        child[up] {node {the}}
        child[up] {node (burger) {burger}}}}};
\end{tikzpicture}
\begin{tikzpicture}[level distance=22pt,sibling distance=24pt]
  \node (arg1) [rectangle,draw=gray]  {$\vur_0$}
      child[up] {node {She}}
      child[up] {node {$\vur_1$}
        child[up] {node {was}}
        child[up] {node {hungry}}
      };
\end{tikzpicture}
\caption{The distributional representations of \example{burger} and \example{hungry} are propagated up the parse tree, clarifying the implicit discourse relation between $\vul_0$ and $\vur_0$.}
\label{fig:upward-example}
\end{subfigure}
\begin{subfigure}{0.48\textwidth}
\centering
\begin{tikzpicture}[level distance=22pt,sibling distance=28pt]
  \node (arg0) {$\vdl_0$}
  child[down,nodraw] {node (bob) {Bob}}
  child[down] {node (vdl1) {$\vdl_1$}
    child[down,nodraw] {node (gave) {gave}}
    child[down] {node (vdl2) {$\vdl_2$}
      child[down] {node (tina) [rectangle,draw=gray] {Tina}}
      child[up,edge from parent/.style={}] {node (vul3) {$\vul_3$} 
        child[up,edge from parent/.style={draw,latex-}] {node {the}}
        child[up,edge from parent/.style={draw,latex-}] {node {burger}}}}};
\draw[->] (bob) -- (vdl1);
\draw[->] (gave) -- (vdl2);
\draw[->] (vul3) -- (tina);
\end{tikzpicture}
\begin{tikzpicture}[level distance=22pt,sibling distance=28pt]
\node (arg1) {$\vdr_0$}
      child[down] {node (she) [rectangle,draw=gray] {She}}
      child[up,edge from parent/.style={}] {node (vur1) {$\vur_1$}
        child[up,edge form parent/.style={draw,latex-}] {node {was}}
        child[up,edge form parent/.style={draw,latex-}] {node {hungry}}
      };
\draw[->] (vur1) -- (she);
\end{tikzpicture}
\caption{Distributional representations for the coreferent mentions \example{Tina} and \example{she} are computed from the parent and sibling nodes.}
\label{fig:downward-example}
\end{subfigure}
\vspace{-2pt}
\caption{Distributional representations are computed through composition over the syntactic parse.}
\label{fig:main-example}
\vspace{-3pt}
\end{figure}
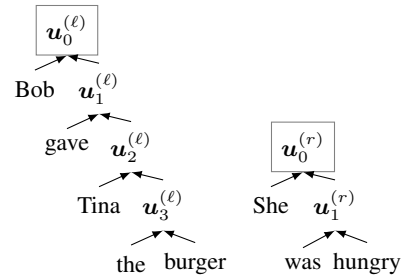
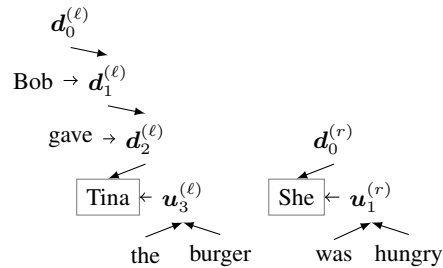

One reason for this poor performance is that predicting implicit discourse relations is a fundamentally semantic task, and the relevant semantics may be difficult to recover from surface level features. For example, consider the implicit discourse relation between the following two sentences (also shown in Figure~\ref{fig:upward-example}): 
\begin{exe}
\ex \example{Bob gave Tina the burger.}\\
\example{She was hungry.}
\label{ex:tina-was-hungry}
\end{exe}
While a connector like \example{because} seems appropriate here, there is little surface information to signal this relationship, unless the model has managed to learn a bilexical relationship between \example{burger} and \example{hungry}. Learning all such relationships from annotated data --- including the relationship of \example{hungry} to \example{knish}, \example{pierogie}, \example{pupusa} etc --- would require far more data than can possibly be annotated.

We address this issue by applying a discriminatively-trained model of compositional distributional semantics to discourse relation classification~\cite{socher2013recursive,baroni2013frege}. The meaning of each discourse argument is represented as a vector~\cite{turney2010frequency}, which is computed through a series of compositional operations over the syntactic parse tree. The discourse relation can then be predicted as a bilinear combination of these vector representations. Both the prediction matrix and the compositional operator are trained in a supervised large-margin framework~\cite{socher2011parsing}, ensuring that the learned compositional operation produces semantic representations that are useful for discourse. We show that when combined with a small number of surface features, this approach outperforms prior work on the classification of implicit discourse relations in the PDTB.

Despite these positive results, we argue that purely vector-based representations are insufficiently expressive to capture discourse relations. To see why, consider what happens if make a tiny change to example~(\ref{ex:tina-was-hungry}):
\begin{exe}
\ex \example{Bob gave Tina the burger.}\\
\example{\textbf{He} was hungry.}
\label{ex:bob-was-hungry}
\end{exe}

\begin{figure}
  \centering
  \includegraphics[width=7.5cm]{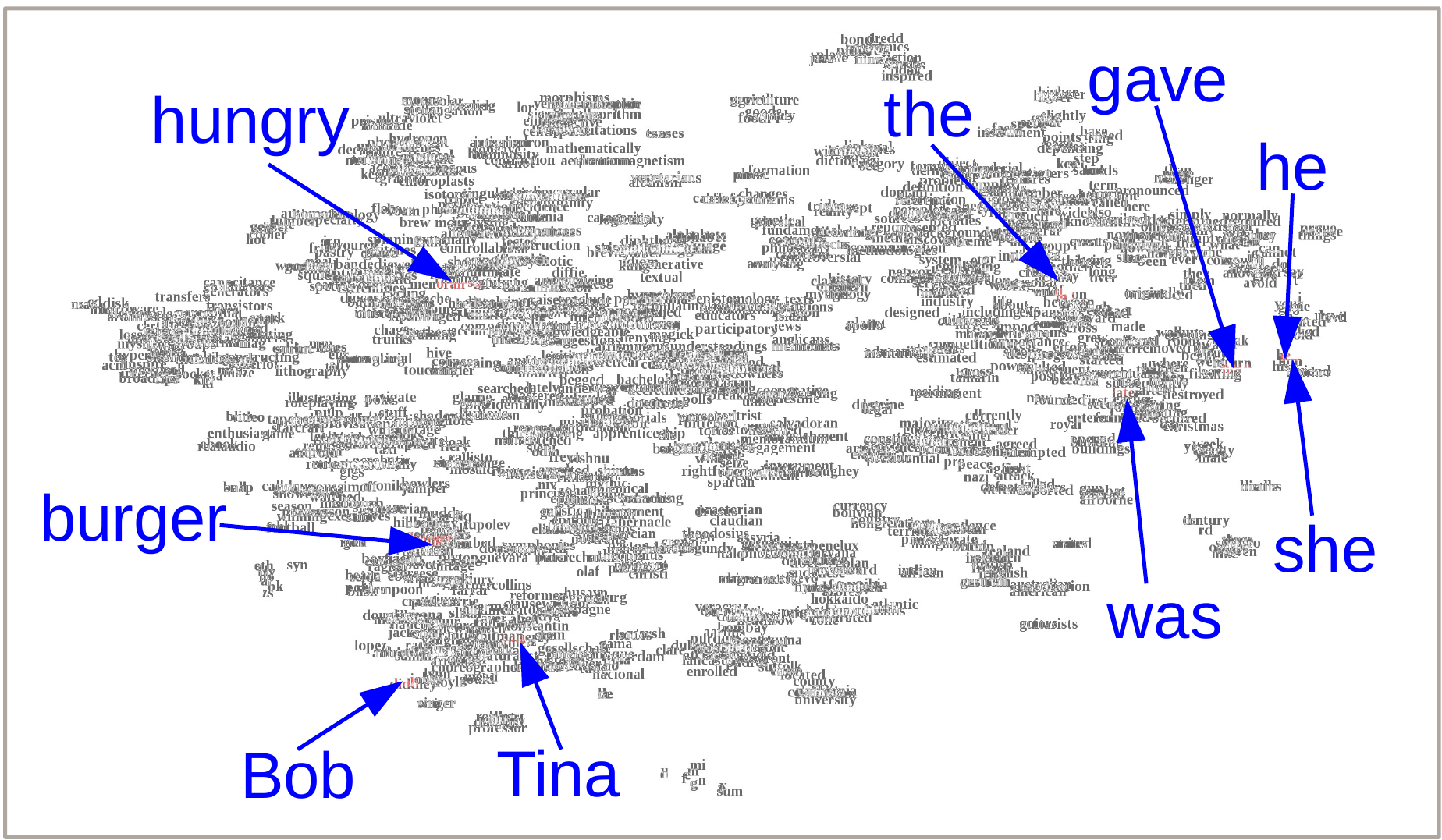}
  \caption{t-SNE visualization~\cite{van2008visualizing} of word representations in the PDTB corpus.}
  \label{fig:latent}
\end{figure}

After changing the subject of the second sentence to Bob, the connective ``\example{because}'' no longer seems appropriate; a contrastive connector like \example{although} is preferred. But despite the radical difference in meaning, the distributional representation of the second sentence will be almost unchanged: the syntactic structure remains identical, and the words \example{he} and \example{she} have very similar word representations (see Figure~\ref{fig:latent}). If we reduce each discourse argument span to a single vector, we cannot possibly capture the ways that discourse relations are signaled by entities and their roles ~\cite{cristea1998veins,louis2010using}. As \newcite{mooney2014semantic} puts it, ``you can't cram the meaning of a whole \%\&!\$\# sentence into a single \$\&!\#* vector!''

We address this issue by computing vector representations not only for each discourse argument, but also for each coreferent entity mention. These representations are meant to capture the \textbf{role} played by the entity in the text, and so they must take the entire span of text into account. We compute entity-role representations using a novel feed-forward compositional model, which combines ``upward'' and ``downward'' passes through the syntactic structure, shown in Figure~\ref{fig:downward-example}. In the example, the downward representations for \example{Tina} and \example{she} are computed from a combination of the parent and sibling nodes in the binarized parse tree. Representations for these coreferent mentions are then combined in a bilinear product, and help to predict the implicit discourse relation. In example~(\ref{ex:bob-was-hungry}), we resolve \example{he} to \example{Bob}, and combine their vector representations instead, yielding a different prediction about the discourse relation.

Our overall approach achieves a 3\% improvement in accuracy over the best previous work~\cite{lin2009recognizing} on multiclass discourse relation classification, and also outperforms more recent work on binary classification. The novel entity-augmented distributional representation improves accuracy over the ``upward'' compositional model, showing the importance of representing the meaning of coreferent entity mentions.

\section{Entity augmented distributional semantics}
\label{sec:model}

We now formally define our approach to entity-augmented distributional semantics, using the notation shown in Table~\ref{tab:notation}. For clarity of exposition, we focus on discourse relations between pairs of sentences. The extension to non-sentence arguments is discussed in Section~\ref{sec:implement}.


\begin{table}
  \centering
  {\small
    \begin{tabular}{lp{5cm}}
      \toprule
      Notation & Explanation\\
      \midrule
      $\ell(i), r(i)$ & left and right children of $i$\\
      $\rho(i), s(i)$ & parent and sibling of $i$\\
      $\alignset{m}{n}$ & set of aligned entities between arguments $m$ and $n$ \\
      $\set{Y}$ & set of discourse relations\\
      $\true{y}$ & ground truth relation\\
      $\psi(y)$ & decision function\\
      $\vu$ & upward vector\\
      $\vd$ & downward vector\\
      $\uclass{y}$ & classification parameter associated with upward vectors\\
      $\dclass{y}$ & classification parameter associated with downward vectors\\
      $\ucomp$ & composition operator in upward composition procedure \\
      $\dcomp$ & composition operator in downward composition procedure \\
      $\mathcal{L}(\vth)$ & objective function \\
      \bottomrule
    \end{tabular}
  }
  \caption{Table of notation}
  \label{tab:notation}
\end{table}

\subsection{Upward pass: argument semantics}
\label{sec:upward}

Distributional representations for discourse arguments are computed in a feed-forward ``upward'' pass: each non-terminal in the binarized syntactic parse tree has a $K$-dimensional distributional representation that is computed from the distributional representations of its children, bottoming out in pre-trained representations of individual words. 

We follow the Recursive Neural Network (RNN) model of \newcite{socher2011parsing}. For a given parent node $i$, we denote the left child as $\lchild{i}$, and the right child as $\rchild{i}$; we compose their representations to obtain, 
\begin{equation}
  \label{eq:upcomp}
  \vu_i = \mytanh{\ucomp [\uvec{\lchild{i}}; \uvec{\rchild{i}}]},
\end{equation}
where $\mytanh{\cdot}$ is the element-wise hyperbolic tangent function~\cite{pascanu2012difficulty}, and $\ucomp\in\mathbb{R}^{K\times 2K}$ is the upward composition matrix. We apply this compositional procedure from the bottom up, ultimately obtaining the argument-level representation $\vu_0$. The base case is found at the leaves of the tree, which are set equal to pre-trained word vector representations. For example, in the second sentence of Figure~\ref{fig:main-example}, we combine the word representations of \example{was} and \example{hungry} to obtain $\vur_1$, and then combine $\vur_1$ with the word representation of \example{she} to obtain $\vur_0$. Note that the upward pass is feedforward, meaning that there are no cycles and all nodes can be computed in linear time.

\subsection{Downward pass: entity semantics}
\label{sec:downward}

As seen in the contrast between Examples~\ref{ex:tina-was-hungry} and~\ref{ex:bob-was-hungry}, a model that uses a single vector representation for each discourse argument would find little to distinguish between \example{she was hungry} and \example{he was hungry}. It would therefore almost certainly fail to identify the correct discourse relation for at least one of these cases, which requires tracking the roles played by the entities that are coreferent in each pair of sentences. To address this issue, we augment the representation of each argument with additional vectors, representing the semantics of the role played by each coreferent entity in each argument. For example, in~(\ref{ex:tina-was-hungry}a), Tina got the burger, and in~(\ref{ex:tina-was-hungry}b), she was hungry. Rather than represent this information in a logical form --- which would require robust parsing to a logical representation --- we represent it through additional distributional vectors.

The role of a constituent $i$ can be viewed as a combination of information from two neighboring nodes in the parse tree: its parent $\rho(i)$, and its sibling $s(i)$. We can make a downward pass, computing the downward vector $\vd_i$ from the downward vector of the parent $\vd_{\rho(i)}$, and the \textbf{upward} vector of the sibling $\vu_{s(i)}$:
\begin{equation}
\label{eq:dcomp}
\vd_i = \mytanh{\dcomp [\dvec{\rho(i)}; \uvec{s(i)}]},
\end{equation}
where $\dcomp\in\mathbb{R}^{K\times 2K}$ is the downward composition matrix. The base case of this recursive procedure occurs at the root of the parse tree, which is set equal to the upward representation, $\vd_0 \triangleq \vu_0$. This procedure is illustrated in Figure~\ref{fig:downward-example}: for \example{Tina}, the parent node is $\vdl_2$, and the sibling is $\vul_3$. 

The up-down compositional algorithm is designed to maintain the \emph{feedforward} nature of the neural network, so that we can efficiently compute all nodes without iterating.  Each downward node $\vd_i$ influences only other downward nodes $\vd_j$ where $j>i$, meaning that the downward pass is feedforward. The upward node is also feedforward: each upward node $\vu_i$ influences only other upward nodes $\vu_j$ where $j < i$. Since the upward and downward passes are each feedforward, and the downward nodes do not influence any upward nodes, the combined up-down network is also feedforward. This ensures that we can efficiently compute all $\vu_i$ and $\vd_i$ in time that is linear in the length of the input.

\paragraph{Connection to the inside-outside algorithm}
In the inside-outside algorithm for computing marginal probabilities in a probabilistic context-free grammar~\cite{lari1990estimation}, the inside scores are constructed in a bottom-up fashion, like our upward nodes; the outside score for node $i$ is constructed from a product of the outside score of the parent $\rho(i)$ and the inside score of the sibling $s(i)$, like our downward nodes. The standard inside-outside algorithm sums over all possible parse trees, but since the parse tree is observed in our case, a closer analogy would be to the constrained version of the inside-outside algorithm for latent variable grammars~\cite{petrov2006learning}. 
\newcite{cohen2014spectral} describe a tensor formulation of the constrained inside-outside algorithm; similarly, we could compute the downward vectors by a tensor contraction of the parent and sibling vectors~\cite{smolensky1990tensor,socher2013reasoning}. However, this would involve $K^3$ parameters, rather than the $K^2$ parameters in our matrix-vector composition.

\section{Predicting discourse relations}
To predict the discourse relation between an argument pair $(m,n)$, the decision function is a sum of bilinear products,

\begin{equation}
  \label{eq:decision}
  {
    \begin{split}
      \psi(y) &= \trans{(\vum_0)} \uclass{y} \vun_{0} \\
      &+ \sum_{i,j \in \set{A}(m,n)} \trans{(\vdm_i)}\dclass{y} \vdn_j + b_y,
    \end{split}
  }
\end{equation}
where $\uclass{y} \in \mathbb{R}^{K\times K}$ and $\dclass{y} \in \mathbb{R}^{K\times K}$ are the classification parameters for relation $y$. A scalar $b_{y}$ is used as the bias term for relation $y$, and $\set{A}(m,n)$ is the set of coreferent entity mentions shared among the argument pair $(m,n)$. The decision value $\psi(y)$ of relation $y$ is therefore based on the upward vectors at the root, $\vum_{0}$ and $\vun_{0}$, as well as on the downward vectors for each pair of aligned entity mentions. For the cases where there are no coreferent entity mentions between two sentences, $\set{A}(m,n) = \varnothing$, the classification model considers only the upward vectors at the root.

To avoid overfitting, we apply a low-dimensional approximation to each $\uclass{y}$,
\begin{equation}
  \label{eq:factor-a}
  {
    \begin{split}
      \uclass{y}=\vec{a}_{y,1}\trans{\vec{a}_{y,2}} + \text{diag}(\vec{a}_{y,3}).    
    \end{split}
  }
\end{equation}
The same approximation is also applied to each $\dclass{y}$, reducing the number of classification parameters from $2 \times \#|\set{Y}|\times K^2$ to $2\times \#|\set{Y}| \times 3K$.

\paragraph{Surface features}
Prior work has identified a number of useful surface-level features~\cite{lin2009recognizing}, and the classification model can easily be extended to include them. Defining $\fvec{(m,n)}$ as the vector of surface features extracted from the argument pair $(m,n)$, the corresponding decision function is modified as,
\begin{equation}
  \label{eq:decision-feat}
  {\small
    \begin{split}
    \psi(y) &= \trans{(\vum_0)} \uclass{y} \vun_{0}
     + \sum_{i,j \in \set{A}(m,n)} 
    \trans{(\vdm_i)}\dclass{y} \vdn_j\\
    &  + \trans{\fclass{y}}\fvec{(m,n)} + b_y,
    \end{split}
  }
\end{equation}
where $\fclass{y}$ is the classification weight on surface features for relation $y$. We describe these features in Section~\ref{sec:implement}.

\section{Large-margin learning framework}
There are two sets of parameters to be learned: the classification parameters $\vth_{class}=\{\uclass{y},\dclass{y},\fclass{y}, b_y\}_{y\in\set{Y}}$, and the composition parameters $\vth_{comp}=\{\ucomp, \dcomp\}$. We use pre-trained word representations, and do not update them. While prior work shows that it can be advantageous to retrain word representations for discourse analysis~\cite{ji2014representation}, our preliminary experiments found that updating the word representations led to serious overfitting in this model.

Following~\newcite{socher2011parsing}, we define a large margin objective, and use backpropagation to learn all parameters of the network jointly~\cite{goller1996learning}. Learning is performed using stochastic gradient descent~\cite{bottou1998online}, so we present the learning problem for a single argument pair $(m,n)$ with the gold discourse relation $\true{y}$. The objective function for this training example is a regularized hinge loss,
\begin{equation}
  \label{eq:objective}
  {\small
    \begin{split}
      \mathcal{L}(\vth) = & \sum_{y':y'\not=\true{y}}\max\Big(0, 1 - \psi(\true{y}) + \psi(y')\Big) + \lambda ||\vth||_2^2
    \end{split}
  }
\end{equation}
where $\vth=\vth_{class}\cup\vth_{comp}$ is the set of learning parameters. The regularization term $\lambda ||\vth||_2^2$ indicates that the squared values of all parameters are penalized by $\lambda$; this corresponds to penalizing the squared Frobenius norm for the matrix parameters, and the squared Euclidean norm for the vector parameters.

\subsection{Learning the classification parameters}
In Equation~\ref{eq:objective}, $\mathcal{L}(\vth)=0$, if for every $y'\not=\true{y}$, $\psi(\true{y})-\psi(y') \geq 1$ holds. Otherwise, the loss will be caused by any $y'$, where $y'\not=\true{y}$ and $\psi(\true{y})-\psi(y') < 1$. The gradient for the classification parameters therefore depends on the margin value between ground truth label and all other labels. 
Specifically, taking one component of $\uclass{y}$, $\vec{a}_{y,1}$, as an example, the derivative of the objective for $y=\true{y}$ is
\begin{equation}
  {\small
    \begin{split}
      \dd{\mathcal{L}(\vth)}{\vec{a}_{y^*,1}} &= -\sum_{y':y'\not=\true{y}}\delta_{(\psi(\true{y})-\psi(y') < 1)}\cdot\vum_{0},
    \end{split}
  }
\end{equation}
where $\delta_{(\cdot)}$ is the delta function. The derivative for $y'\not=\true{y}$ is
\begin{equation}
  {\small
    \begin{split}
      \dd{\mathcal{L}(\vth)}{\vec{a}_{y',1}} &= \delta_{(\psi(\true{y})-\psi(y') < 1)}\cdot\vum_{0}
    \end{split}
  }
\end{equation}

During learning, the updating rule for $\uclass{y}$ is 
\begin{equation}
  \label{eq:update-uclass}
  {\small
    \begin{split}
      \uclass{y}&\leftarrow\uclass{y} - \eta (\dd{\mathcal{L}(\vth)}{\uclass{y}} + \lambda\uclass{y})
    \end{split}
  }
\end{equation}
where $\eta$ is the learning rate. 

Similarly, we can obtain the gradient information and updating rules for parameters $\{\dclass{y},\fclass{y},b_y\}_{y\in\set{Y}}$.

\subsection{Learning the composition parameters}
There are two composition matrices $\ucomp$ and $\dcomp$, corresponding to the upward and downward composition procedures respectively. Taking the upward composition parameter $\ucomp$ as an example, the derivative of $\mathcal{L}(\vth)$ with respect to $\ucomp$ is
\begin{equation}
  {\small 
    \begin{split}
      \dd{\mathcal{L}(\vth)}{\ucomp} = & \sum_{y':y'\not=\true{y}}\delta_{(\psi(\true{y})-\psi(y') < 1)}\\
      & \cdot\Big(\dd{\psi(y')}{\ucomp}-\dd{\psi(\true{y})}{\ucomp}\Big)
    \end{split}
  }
\end{equation}
As with the classification parameters, the derivative depends on the margin between $y'$ and $\true{y}$. 

For every $y\in\set{Y}$, we have the unified derivative form,
\begin{equation}
  \label{eq:psi-ucomp}
  {\small
    \begin{split}
      \dd{\psi(y)}{\ucomp} &= \dd{\psi(y)}{\vum_{0}}\dd{\vum_{0}}{\ucomp} 
      + \dd{\psi(y)}{\vun_{0}}\dd{\vun_{0}}{\ucomp}\\
      &+ \sum_{i,j\in\alignset{m}{n}}\dd{\psi(y)}{\vdm_{i}}\dd{\vdm_{i}}{\ucomp}\\
      &+ \sum_{i,j\in\alignset{m}{n}}\dd{\psi(y)}{\vdn_{j}}\dd{\vdn_{j}}{\ucomp},
    \end{split}
  }
\end{equation}
The gradient information of $\ucomp$ also depends on the gradient information of $\psi(y)$ with respect to every downward vector $\dvec{}$, as shown in the last two terms in Equation \ref{eq:psi-ucomp}. This is because the computation of each downward vector $\vd_i$ includes the upward vector of the sibling node, $\vu_{s(i)}$, as shown in Equation~\ref{eq:dcomp}. For an example, see the construction of the downward vectors for \example{Tina} and \example{she} in Figure~\ref{fig:downward-example}.

The partial derivatives of the decision function in Equation~\ref{eq:psi-ucomp} are computed as,
\begin{equation}
  {\small
    \begin{split}
      \dd{\psi(y)}{\vum_{0}} = A_y \vun_0, &~ \dd{\psi(y)}{\vun_{0}} = \trans{A_y} \vum_0,\\
      \dd{\psi(y)}{\vdm_{i}} = B_y \vdn_{j}, &~ \dd{\psi(y)}{\vdn_{i}} = \trans{B_y} \vdm_{j}, \tuple{i,j} \in \set{A}.
    \end{split}
  }
\end{equation}

The partial derivatives of the upward and downward vectors with respect to the upward compositional operator are computed as,
\begin{equation}
  {\small
    \begin{split}
      \dd{\vum_i}{\ucomp} &= \sum_{\vum_{k}\in\set{T}(\vum_i)}\dd{\vum_i}{\vum_k}\trans{(\vum_k)}
    \end{split}
  }
\end{equation}
\vskip -1em
and
\begin{equation}
  {\small
    \begin{split}
    \dd{\vdm_i}{\ucomp} = \sum_{\vum_k\in\set{T}(\vdm_i)}\dd{\vdm_i}{\vum_k}\trans{(\vum_k)},
    \end{split}
  }
\end{equation}
where $\set{T}(\uvec{m})$ is the set of all nodes in the upward composition model that help to generate $\uvec{m}$. For example, in Figure~\ref{fig:upward-example}, the set $\set{T}(\vu^{(\ell)}_2)$ includes $\vul_3$ and the word representations for \example{Tina}, \example{the}, and \example{burger}. The set $\set{T}(\dvec{m,i})$ includes all the \textbf{upward} nodes involved in the downward composition model generating $\vdm_i$. For example, in Figure~\ref{fig:downward-example}, the set $\set{T}(\vdr_{\text{she}})$ includes $\vur_1$ and the word representations for \example{was} and \example{hungry}.

The derivative of the objective with respect to the downward compositional operator $\dcomp$ is computed in a similar fashion, but it depends only on the downward nodes, $\vdm_i$.

\section{Implementation}\label{sec:implement}
Our implementation will be made available online after review. Training on the PDTB takes roughly three hours to converge.\footnote{On Intel(R) Xeon(R) CPU 2.20GHz without parallel computing.} Convergence is faster if the surface feature weights $\vec{\beta}$ are trained separately first. We now describe some additional details of our implementation.

\paragraph{Learning}
During learning, we used AdaGrad \cite{duchi2011adaptive} to tune the learning rate in each iteration. To avoid the exploding gradient problem~\cite{bengio1994learning}, we used the norm clipping trick proposed by \newcite{pascanu2012difficulty}, fixing the norm threshold at $\tau=5.0$.

\paragraph{Parameters}
Our model includes three tunable parameters: the latent dimension $K$ for the distributional representation, the regularization parameter $\lambda$, and the initial learning rate $\eta$. All parameters are selected by randomly selecting a development set of 20\% of the training data. We consider the values $K\in \{20,30,40,50,60\}$ for the latent dimensionality, $\lambda\in \{0.0002,0.002,0.02,0.2\}$ for the regularization (on each training instance), and $\eta \in \{0.01,0.03,0.05,0.09\}$ for the learning rate. We assign separate regularizers and learning rates to the upward composition model, downward composition model, feature model and the classification model with composition vectors.

\paragraph{Initialization}
All the classification parameters are initialized to $\vec{0}$. For the composition parameters, we follow \newcite{bengio2012practical} and initialize $\ucomp$ and $\dcomp$ with uniform random values drawn from the range $[-\sqrt{{6}/{2K}},\sqrt{{6}/{2K}}]$. 

\paragraph{Word representations}
We trained a word2vec model \cite{mikolov2013linguistic} on the PDTB corpus, standardizing the induced representations to zero-mean, unit-variance~\cite{lecun2012efficient}. Experiments with pre-trained GloVe word vector representations~\cite{pennington2014glove} gave broadly similar results.

\paragraph{Syntactic structure}
Our model requires that the syntactic structure for each argument as a binary tree. We run the Stanford parser \cite{klein2003accurate} to obtain constituent parse trees of each sentence in the PDTB, and binarize all resulting parse trees. Argument spans in the Penn Discourse Treebank need not be sentences or syntactic constituents: they can include multiple sentences, non-constituent spans, and even discontinuous spans~\cite{prasad2008penn}. In all cases, we identify the syntactic subtrees within the argument span, and construct a right branching superstructure that unifies them into a tree.

\begin{table}
  \centering
  {\small
    \begin{tabular}{llll}
      \toprule
      Dataset & Annotation & Training (\%) & Test (\%)\\
      \midrule
      1. PDTB & Automatic & 27.4 & 29.1\\
      2. \ontodata & Automatic & 26.2 & 32.3\\
      3. \ontodata & Gold & 40.9 & 49.3\\
      \bottomrule
    \end{tabular}
  }
  \caption{Proportion of relations with coreferent entities, according to automatic coreference resolution and gold coreference annotation.}
  \label{tab:shared}
\end{table}

\paragraph{Coreference}
To extract entities from the PDTB, we ran the Berkeley coreference system \cite{DurrettKlein2013} on each document. For each argument pair, we simply ignore the non-corefential entity mentions. Line 1 in Table~\ref{tab:shared} shows the proportion of the instances with shared entities in the PDTB training and test data. We also consider the intersection of the PDTB with the OntoNotes corpus~\cite{pradhan2007ontonotes}, which contains gold coreference annotations. The intersection \ontodata~contains 597 documents; the statistics for automatic and gold coreference are shown in lines 2 and 3 of Table~\ref{tab:shared}.

\paragraph{Additional features}
We supplement our classification model using additional surface features proposed by~\newcite{lin2009recognizing}. These include four categories: lexical features, constituent parse features, dependency parse features, and contextual features. Following this prior work, we use mutual information to select features in the first three categories, obtaining 500 lexical features, 100 constituent features, and 100 dependency features.



\section{Experiments}
We evaluate our approach on the Penn Discourse Treebank (PDTB)~\cite{prasad2008penn}, which provides a discourse level annotation over the Wall Street Journal corpus. In the PDTB, each discourse relation is annotated between two argument spans. Identifying the argument spans of discourse relations is a challenging task~\cite{lin2014pdtb}, which we do not attempt here; instead, we use gold argument spans, as in most of the prior work on this task. PDTB relations may be \textbf{explicit}, meaning that they are signaled by discourse connectives (e.g., \example{because}); alternatively, they may be \textbf{implicit}, meaning that the connective is absent. \newcite{pitler2008easily} show that most connectives are unambiguous, so we focus on the more challenging problem of classifying implicit discourse relations.

The PDTB provides a three-level hierarchy of discourse relations. The first level consists of four major relation \textbf{classes}: \rela{Temporal, Contingency, Comparison} and \rela{Expansion}. For each class, a second level of \textbf{types} is defined to provide finer semantic distinctions; there are sixteen such relation types. A third level of \textbf{subtypes} is defined for only some types, specifying the semantic contribution of each argument. 

There are two main approaches to evaluating implicit discourse relation classification. Multiclass classification requires identifying the discourse relation from all possible choices. This task was explored by \newcite{lin2009recognizing}, who focus on second-level discourse relations. More recent work has emphasized binary classification, where the goal is to build and evaluate separate ``one-versus-all'' classifiers for each discourse relation~\cite{pitler2009automatic,park2012implicit,biran2013aggregated}. We primarily focus on multiclass classification, because it is more relevant for the ultimate goal of building a PDTB parser; however, to compare with recent prior work, we also evaluate on binary relation classification.


\begin{table*}
  \centering
  {\small
  \begin{tabular}{lllll}
    \toprule
    Model &  +Entity semantics & +Surface features & $K$ & Accuracy(\%) \\
    \midrule
    \emph{Baseline models}\\
    1. Most common class  &   & No   &  & 26.03\\
    2. Additive word representations  &  & No & 50 & 28.73\\[0.5em]
    \emph{Prior work}\\
    3. \cite{lin2009recognizing}  &   & Yes  &  & 40.2\\[0.5em]
    \emph{Our work}\\
    4. Surface feature model  &   & Yes  &  & 39.69\\[0.5em]
    5. \modelname   & No & No & 50 & 36.98 \\
    6. \modelname   & Yes & No & 50 & 37.63\\[0.3em]
    7. \modelname   & No & Yes  & 50 & 42.53$^\dag$\\
    8. \modelname   & Yes & Yes  & 50 & {\bf 43.56}$^\ast$\\
    \bottomrule
    \multicolumn{4}{l}{$^\ast$ signficantly better than \cite{lin2009recognizing} with $p<0.05$}\\
    \multicolumn{4}{l}{$^\dag$ signficantly better than line 4 with $p<0.05$}
  \end{tabular}
  }
  \caption{Experimental results on multiclass classification of level-2 discourse relations. The results of \newcite{lin2009recognizing} are shown in line 3; the results for our reimplementation of this system are shown in line 4.}
  \label{tab:results-pdtb}
\end{table*}

\subsection{Multiclass classification}
\label{sec:multiclass}
Our main evaluation involves predicting the correct discourse relation for each argument pair, from among the second-level relation types. Following \newcite{lin2009recognizing}, we exclude five relation types that are especially rare: \rela{Condition, Pragmatic Condition, Pragmatic Contrast, Pragmatic Concession} and \rela{Expression}. In addition, about 2\% of the implicit relations in the PDTB are annotated with more than one type. During training, each argument pair that is annotated with two relation types is considered as two training instances, each with one relation type. During testing, if the classifier assigns either of the two types, it is considered to be correct.

\subsubsection{Baseline and competitive systems}
\begin{description}[topsep=0pt,itemsep=-1ex,partopsep=1ex,parsep=1ex]
\item[Most common class] The most common class is \rela{Cause}, accounting for 26.03\% of the implicit discourse relations in the PDTB test set.
\item[Additive word representations] \newcite{blacoe2012comparison} show that simply adding word vectors can perform surprisingly well at assessing the meaning of short phrases. In this baseline, we represent each argument as a sum of its word representations, and estimate a bilinear prediction matrix.
\item[\newcite{lin2009recognizing}] To our knowledge, the best published accuracy on multiclass classification of second-level implicit discourse relations is from \newcite{lin2009recognizing}, who apply feature selection to obtain a set of lexical and syntactic features over the arguments.
\item[Surface feature model] We re-implement the system of \newcite{lin2009recognizing}, enabling a more precise comparison. The major difference is that we apply our online learning framework, rather than a batch classification algorithm.
\item[Compositional] Finally, we report results for the method described in this paper. Since it is a \textbf{dis}tributional \textbf{co}mpositional approach to \textbf{disco}urse relations, we name it \modelname.
\end{description}

\subsubsection{Results}
Table~\ref{tab:results-pdtb} presents results for multiclass identification of second-level PDTB relations. As shown in lines 7 and 8, \modelname~outperforms both baseline systems and the prior state-of-the-art (line 3). The strongest performance is obtained by including the entity distributional semantics, with a 3.4\% improvement over the accuracy reported by \newcite{lin2009recognizing} ($p < .05$). The improvement over our reimplementation of this work is even greater, which shows how the distributional representation provides additional value over the surface features. Because we have reimplemented this system, we can observe individual predictions, and can therefore use the more sensitive sign test for statistical significance. This test shows that even without entity semantics, \modelname~significantly outperforms the surface feature model ($p < .05$). 

\begin{figure}
  \centering
  \includegraphics[width=7.5cm]{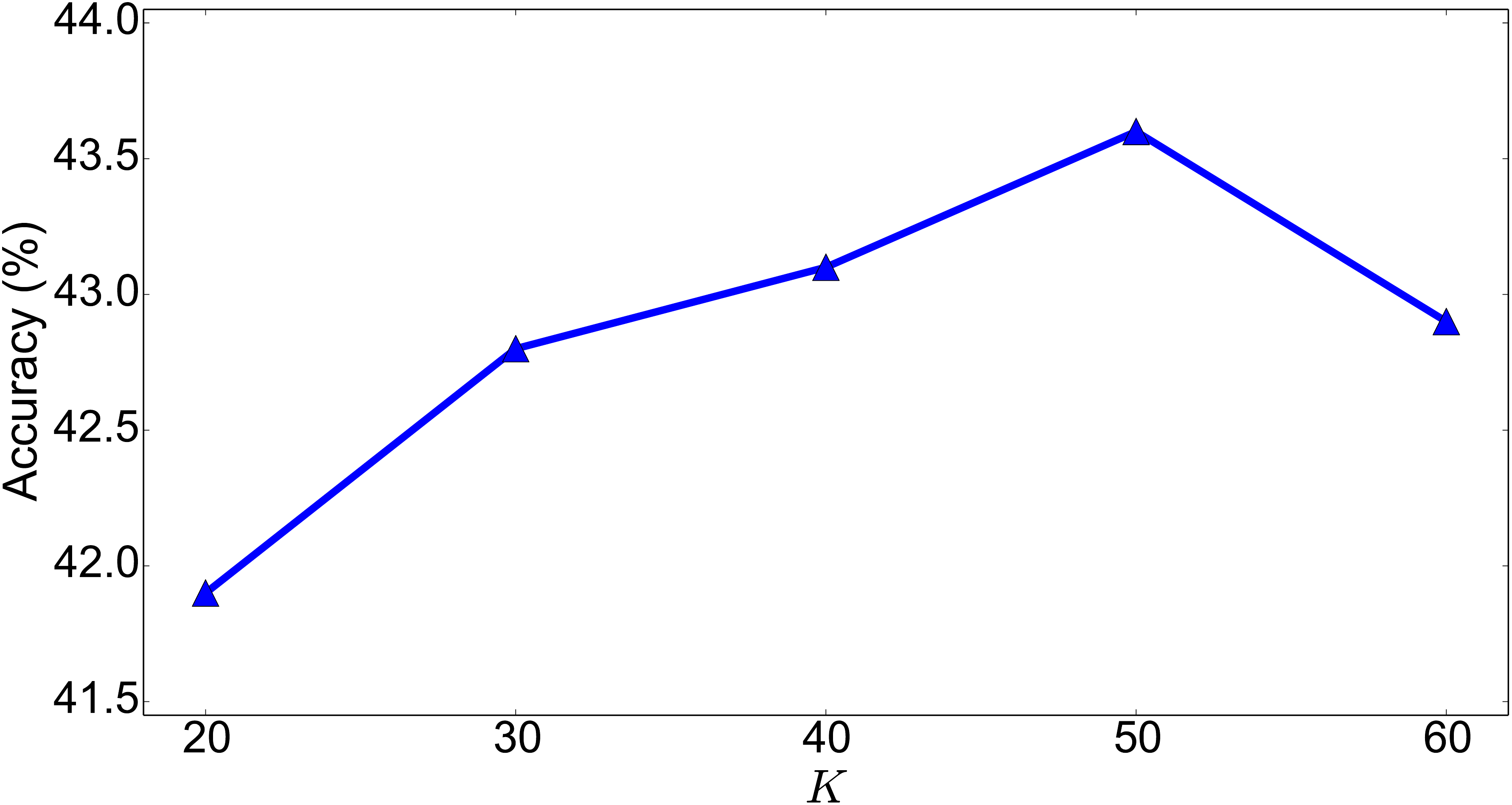}
  \caption{The performance of \modelname~(full model), over different latent dimensions $K$.}
  \label{fig:accuracy}
\end{figure}

The latent dimension $K$ is chosen from a development set (see Section~\ref{sec:implement}). Test set performance for each setting of $K$ is shown in Figure~\ref{fig:accuracy}, with accuracies in a narrow range between 41.9\% and 43.6\%.

\subsubsection{Coreference}
\label{sec:coref}
The contribution of entity semantics is shown in Table~\ref{tab:results-pdtb} by the accuracy differences between lines 5 and 6, and between lines 7 and 8. On the subset of relations in which the arguments share at least one coreferent entity, the difference is substantially larger: the accuracy of \modelname~is 44.9\% with entity semantics, and 42.2\% without. Considering that only 29.1\% of the relations in the PDTB test set include shared entities, it therefore seems likely that a more sensitive coreference system could yield further improvements for the entity-semantics model. Indeed, gold coreference annotation on the intersection between the PDTB and the OntoNotes corpus shows that 40-50\% of discourse relations involve coreferent entities (Table~\ref{tab:shared}). Evaluating our model on just this intersection, we find that the inclusion of entity semantics yields an improvement in accuracy from 37.1\% to 38.8\%.

\subsection{Binary classification}
\label{sec:binary}
Much of the recent work in PDTB relation detection has focused on binary classification, building and evaluating separate one-versus-all classifiers for each relation type~\cite{pitler2009automatic,park2012implicit,biran2013aggregated}.
This work has focused on recognition of the four \emph{first}-level relations, grouping \rela{EntRel} with the \rela{Expansion} relation. We follow this evaluation approach as closely as possible, using sections 2-20 of the PDTB as a training set, sections 0-1 as a development set for parameter tuning, and sections 21-22 for testing.

\subsubsection{Classification method}
We apply \modelname~with the downward composition procedure and the same surface features listed in Section~\ref{sec:implement}; this corresponds to the system reported in line 8 of Table~\ref{tab:results-pdtb}. However, instead of employing a multiclass classifier for all four relations, we train four binary classifiers, one for each first-level discourse relation. We optimize the hyperparameters $K,\lambda,\eta$ separately for each classifier (see Section~\ref{sec:implement} for details), by performing a grid search to optimize the F-measure on the development data. Following 
 \newcite{pitler2009automatic}, we obtain a balanced training set by resampling training instances in each class until the number of positive and negative instances are equal.

\subsubsection{Competitive systems}
We compare our model with the published results from several competitive systems. Since we are comparing with previously published results, we focus on systems which use the predominant training / test split, with sections 2-20 for training and 21-22 for testing. This means we cannot compare with recent work from \newcite{li2014reducing}, who use sections 20-24 for testing.
\begin{description}[topsep=0pt,itemsep=-1ex,partopsep=1ex,parsep=1ex]
\item[\newcite{pitler2009automatic}] present a classification model using linguistically-informed features, such as polarity tags and Levin verb classes.
\item[\newcite{zhou2010predicting}] predict discourse connective words, and then use these predicted connectives as features in a downstream model to predict relations.
\item[\newcite{park2012implicit}] showed that the performance on each relation can be improved by selecting a locally-optimal feature set.
\item[\newcite{biran2013aggregated}] reweight word pair features using distributional statistics from the Gigaword corpus, obtaining denser aggregated score features.
\end{description}

\subsubsection{Experimental results}


\begin{table*}
  \centering
  {\small 
  \begin{tabular}{lllllllll}
    \toprule
    & \multicolumn{2}{l}{\rela{Comparison}} & \multicolumn{2}{l}{\rela{Contingency}} & \multicolumn{2}{l}{\rela{Expansion}} & \multicolumn{2}{l}{\rela{Temporal}} \\
    & F1 & Acc & F1 & Acc & F1 & Acc & F1 & Acc\\
    \midrule
    {\em Competitive systems}\\
    1. \cite{pitler2009automatic} & 21.96 & 56.59 & 47.13 & 67.30 & 76.42 & 63.62 & 16.76 & 63.49\\
    2. \cite{zhou2010predicting} & 31.79 & 58.22 & 47.16 & 48.96 & 70.11 & 54.54 & 20.30 & 55.48\\
    3. \cite{park2012implicit} & 31.32 & {\bf 74.66} & 49.82 & 72.09 & 79.22 & 69.14 & 26.57 & 79.32\\
    4. \cite{biran2013aggregated} & 25.40 & 63.36 & 46.94 & 68.09 & 75.87 & 62.84 & 20.23 & 68.35\\[0.2em]
    {\em Our work}\\
    5. \modelname & {\bf 35.84} & 68.45 & {\bf 51.39} & {\bf 74.08} & {\bf 79.91} & {\bf 69.47} & {\bf 26.91} & {\bf 86.41}\\
    \bottomrule
  \end{tabular}
  }
  \caption{Evaluation on the first-level discourse relation identification. The results of the competitive systems are reprinted.}
  \label{tab:level-one}
\end{table*}

Table~\ref{tab:level-one} presents the performance of the \modelname~model and the published results of competitive systems. Our model achieves the best results on most metrics, achieving F-measure improvements of 4.52\% on \rela{Comparison}, 1.57\% on \rela{Contingency}, 0.69\% on \rela{Expansion}, and 0.34\% on \rela{Temporal}. These results are attained without performing per-relation feature selection, as in prior work.

\section{Related Work}\label{sec:related}
This paper draws mainly on previous work in discourse relation detection and compositional distributional semantics.

\subsection{Discourse relations}
Many models of discourse structure focus on relations between spans of text~\cite{knott1996data}, including rhetorical structure theory (RST; Mann and Thompson, 1988)\nocite{mann1988rhetorical}, lexical tree-adjoining grammar for discourse (D-LTAG; Webber, 2004)\nocite{webber2004dltag}, and even centering theory~\cite{grosz1995centering}, which posits relations such as \rela{continuation} and \rela{smooth shift} between adjacent spans. Consequently, the automatic identification of discourse relations has long been considered a key component of discourse parsing~\cite{marcu1999decision}. 

We work within the D-LTAG framework, as annotated in the Penn Discourse Treebank (PDTB; Prasad et al., 2008)\nocite{prasad2008penn}, with the task of identifying \emph{implicit} discourse relations. The seminal work in this task is from \newcite{pitler2009automatic} and \newcite{lin2009recognizing}. \newcite{pitler2009automatic} focus on lexical features, including linguistically motivated word groupings such as Levin verb classes and polarity tags. \newcite{lin2009recognizing} identify four different feature categories, based on the raw text, the context, and syntactic parse trees; the same feature sets are used in later work on end-to-end discourse parsing~\cite{lin2014pdtb}, which also includes components for identifying argument spans. Subsequent research has explored feature selection~\cite{park2012implicit,lin2014pdtb}, as well as combating feature sparsity by aggregating features~\cite{biran2013aggregated}.
Our model includes surface features that are based on a reimplementation of the work of \newcite{lin2009recognizing}, because they also undertake the task of multiclass relation classification; however, the techniques introduced in more recent research may also be applicable and complementary to the distributional representation that constitutes the central contribution of this paper; if so, applying these techniques could further improve performance.

Our contribution of entity-augmented distributional semantics is motivated by the intuition that entities play a central role in discourse structure. Centering theory draws heavily on referring expressions to entities over the discourse~\cite{grosz1995centering,barzilay2008modeling}; similar ideas have been extended to rhetorical structure theory~\cite{corston1998beyond,cristea1998veins}. In the specific case of identification of implicit PDTB relations, \newcite{louis2010using} explore a number of entity-based features, including grammatical role, syntactic realization, and information status. Despite the solid linguistic foundation for these features, they are shown to contribute little in comparison with more traditional word-pair features. This suggests that syntax and information status may not be enough, and that it is crucial to capture \emph{semantics} of each entity's role in the discourse. Our approach does this by propagating distributional semantics from throughout the sentence into the entity span, using our up-down compositional procedure.

\subsection{Compositional distributional semantics}
Distributional semantics begins with the hypothesis that words and phrases that tend to appear in the same contexts have the same meaning~\cite{firth1957papers}. The current renaissance of interest in distributional semantics can be attributed in part to the application of discriminative techniques, which emphasize \emph{predictive} models~\cite{bengio2006neural,baroni2014don}, rather than context-counting and matrix factorization~\cite{LSA,turney2010frequency}. In addition, recent work has made practical the idea of propagating distributional information through linguistic structures~\cite{smolensky1990tensor,collobert2011natural}. In such models, the distributional representations and compositional operators can be fine-tuned by backpropagating supervision from task-specific labels, enabling accurate and fast models for a wide range of language technologies~\cite{socher2011parsing,socher2013recursive,chen2014fast}.

The application of distributional semantics to discourse includes the use of latent semantic analysis for text segmentation~\cite{choi2001latent} and coherence assessment~\cite{foltz1998measurement}, as well as paraphrase detection by the factorization of matrices of distributional counts~\cite{kauchak2006paraphrasing,mihalcea2006corpus}. These approaches essentially compute a distributional representation in advance, and then use it alongside other features. In contrast, our approach follows more recent work in which the distributional representation is driven by supervision from discourse annotations. For example, \newcite{ji2014representation} show that RST parsing can be performed by learning task-specific word representations, which perform considerably better than generic word2vec representations~\cite{mikolov2013linguistic}. \newcite{li2014recursive} propose a recurrent neural network approach to RST parsing, which is similar to the upward pass in our model. However, prior work has not applied these ideas to the classification of implicit relations in the PDTB, and does not consider the role of entities. As we argue in the introduction, a single vector representation is insufficiently expressive, because it obliterates the entity chains that help to tie discourse together.

More generally, our entity-augmented distributional representation can be viewed in the context of recent literature on combining distributional and formal semantics: by representing entities, we are taking a small step away from purely distributional representations, and towards more traditional logical representations of meaning. In this sense, our approach is ``bottom-up'', as we try to add a small amount of logical formalism to distributional representations; other approaches are ``top-down'', softening purely logical representations by using distributional clustering~\cite{poon2009unsupervised,lewis2013combined} or Bayesian non-parametrics~\cite{titov2011bayesian} to obtain types for entities and relations. Still more ambitious would be to implement logical semantics within a distributional compositional framework~\cite{clark2011mathematical,grefenstette2013towards}. At present, these combinations of logical and distributional semantics have been explored only at the sentence level. In generalizing such approaches to multi-sentence discourse, we argue that it will not be sufficient to compute distributional representations of sentences: a multitude of other elements, such as entities, will also have to represented.

\section{Conclusion}
Discourse relations are determined by the meaning of their arguments, and progress on discourse parsing therefore requires computing representations of the argument semantics. We present a compositional method for inducing distributional representations not only of discourse arguments, but also of the entities that thread through the discourse. In this approach, semantic composition is applied up the syntactic parse tree to induce the argument-level representation, and then down the parse tree to induce representations of entity spans. Discourse arguments can then be compared in terms of their overall distributional representation, as well as by the representations of coreferent entity mentions. This enables the compositional operators to be learned by backpropagation from discourse annotations. This approach outperforms previous work on classification of implicit discourse relations in the Penn Discourse Treebank. Future work may consider joint models of discourse structure and coreference, as well as representations for other discourse elements, such as event coreference and shallow semantics.

\bibliographystyle{acl}
\bibliography{ref,cite-strings,cites,cite-definitions}
\end{document}